\documentclass[11pt, a4paper,logo,  copyright, nonumbering]{paddleocr}
\usepackage[numbers]{natbib}
\usepackage{dblfloatfix}
\usepackage{ulem}
\usepackage{float}
\usepackage{tocloft}
\usepackage{caption}
\usepackage{dramatist}
\usepackage{xspace}
\usepackage{array}
\usepackage{pifont} %
\usepackage{multirow}
\usepackage{geometry}

\usepackage{makecell}
\usepackage{tcolorbox}
\usepackage{xltabular}
\usepackage{titlesec} 
\usepackage{longtable}
\usepackage[hang, flushmargin]{footmisc}
\usepackage{booktabs} 
\usepackage{xcolor}   
\usepackage[table]{xcolor}
\usepackage{algorithm}
\usepackage{algorithmic}
\definecolor{lightpink}{RGB}{255, 182, 193}
\definecolor{iccvblue}{rgb}{0.21,0.49,0.74}

\hypersetup{
    colorlinks=true,
    linkcolor=blue,
    citecolor=blue,
    filecolor=blue,
    urlcolor=blue,
    pdfpagemode=UseNone,
    pdfstartview=FitH
}
\usepackage{threeparttable} 
\usepackage{soul}
\usepackage{amsfonts}
\usepackage{amsmath}
\usepackage{amssymb}
\usepackage{lineno}
\usepackage{multirow}
\usepackage{adjustbox}

\usepackage{CJKutf8}
\usepackage{subcaption}
\usepackage{setspace}

\usepackage{dsfont}
\usepackage{array} %
\usepackage{tabularx} %
\usepackage{xcolor} %
\usepackage{tabularx}
\usepackage{booktabs}

\usepackage{lipsum}  %
\usepackage{multicol} %
\usepackage{makecell} 
\usepackage{listings}
\usepackage{xcolor}

\usepackage{tablefootnote}
\usepackage{hyperref}

\lstdefinelanguage{json}{
    basicstyle=\ttfamily\footnotesize,
    numbers=left,
    numberstyle=\tiny\color{gray},
    stepnumber=1,
    numbersep=8pt,
    showstringspaces=false,
    breaklines=true,
    frame=lines,
    backgroundcolor=\color{gray!10},
    morestring=[b]",
    literate=
     *{0}{{{\color{black}0}}}{1}
      {1}{{{\color{black}1}}}{1}
      {2}{{{\color{black}2}}}{1}
      {3}{{{\color{black}3}}}{1}
      {4}{{{\color{black}4}}}{1}
      {5}{{{\color{black}5}}}{1}
      {6}{{{\color{black}6}}}{1}
      {7}{{{\color{black}7}}}{1}
      {8}{{{\color{black}8}}}{1}
      {9}{{{\color{black}9}}}{1}
}

\makeatletter
\def\@BTrule[#1]{%
  \ifx\longtable\undefined
    \let\@BTswitch\@BTnormal
  \else\ifx\hline\LT@hline
    \nobreak
    \let\@BTswitch\@BLTrule
  \else
     \let\@BTswitch\@BTnormal
  \fi\fi
  \global\@thisrulewidth=#1\relax
  \ifnum\@thisruleclass=\tw@\vskip\@aboverulesep\else
  \ifnum\@lastruleclass=\z@\vskip\@aboverulesep\else
  \ifnum\@lastruleclass=\@ne\vskip\doublerulesep\fi\fi\fi
  \@BTswitch}
\makeatother

\addto\extrasenglish{
}

 {\begin{list}{}%
         {\setlength{\leftmargin}{#1}}%
         \item[]%
 }
 {\end{list}}

\reportnumber{001} %

\renewcommand{\today}{}

\title{\centering PaddleOCR-VL-1.6: Expanding the Frontier of Document Parsing with Under-Optimized Region Refinement and Progressive Post-Training}

\author[*]{
\small
Zelun Zhang, Hongen Liu, Suyin Liang, Yubo Zhang, Yiqing Xiang, 
\vspace{-0.4cm}
\\
\small
Jiaxuan Liu, Ting Sun, Manhui Lin, Yue Zhang, Changda Zhou, 
\\
\small
Tingquan Gao, Cheng Cui$^{\dagger}$, Yi Liu, Dianhai Yu, Yanjun Ma
\vspace{0.2cm}
\\
\small
\textbf{PaddlePaddle Team, Baidu Inc.} \\
\small
$^{\dagger}$Project Leader
\vspace{0.2cm}
  \\
  
  {\small
  \raggedright{  
  \small
  \hspace{11.0em}  
  \includegraphics[height=1.0em]{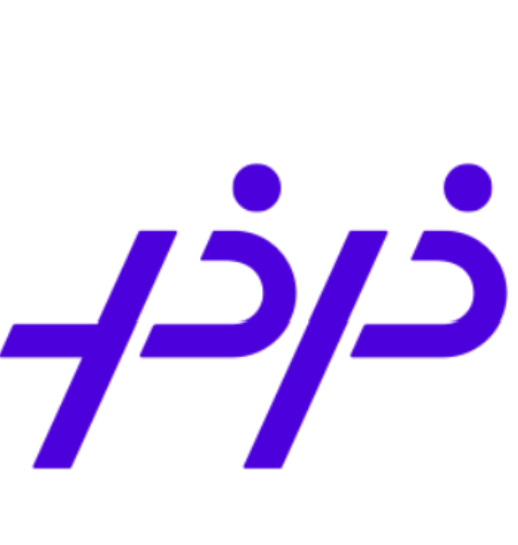} \textbf{Official Website}: \url{https://www.paddleocr.com} \\
  \hspace{-1.2em}  
  \small
  \hspace{6.55em}  
  \includegraphics[height=0.9em]{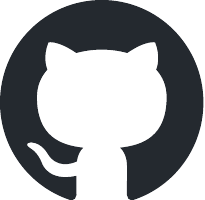} \textbf{Source Code}: \url{https://github.com/PaddlePaddle/PaddleOCR} \\
  \hspace{-1.2em}  
  \small
  \hspace{6.9em}  
  \includegraphics[height=1.0em]{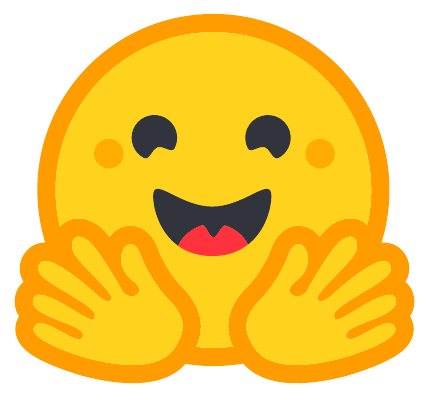} \textbf{Models}: \url{https://huggingface.co/PaddlePaddle} \\
  }
  }
  
}

\renewcommand{\phi}{\varphi}

\renewcommand{\geq}{\geqslant}

\renewcommand{\epsilon}{\varepsilon}
\renewcommand{\imath}{\mathrm{i}}

\newlength{\restsubwidth}
\newlength{\restsubheight}
\newlength{\restsubmoreheight}
\newcommand{\rest}[2]{%
        \settowidth{\restsubwidth}{\ensuremath{#2}}
        \settoheight{\restsubheight}{\ensuremath{{}_{#2}}}
        \ensuremath{{#1\hskip 0.5pt}_{\vrule\kern2pt\parbox[b][%
        4pt][b]{\the\restsubwidth}{%
                        \ensuremath{{}_{#2}}}}}
        }

\begin{abstract}

\vspace{-0.5cm} 

We introduce PaddleOCR-VL-1.6, an upgraded compact document parsing model built upon PaddleOCR-VL-1.5. Although PaddleOCR-VL-1.5 establishes a strong 0.9B baseline, its remaining errors concentrate in under-optimized regions where model behavior is unstable, data coverage is sparse, or supervision is unreliable. Rather than expanding the training corpus indiscriminately, PaddleOCR-VL-1.6 introduces a region-aware data optimization framework that identifies weak regions from the previous model, applies targeted enhancement to these regions, and improves the reliability of supervision signals. It further adopts a progressive post-training recipe based on curated data selection and reinforcement learning, pushing model performance to a higher level through staged optimization. PaddleOCR-VL-1.6 achieves a new state-of-the-art score of 96.33\% on OmniDocBench v1.6, demonstrates strong competitiveness against top-tier VLMs, and provides a practical post-training recipe for the PaddleOCR-VL series.

\end{abstract}

\begin{document}

\newgeometry{
  top=1.8cm,
  bottom=1.8cm,
  headsep=0.1cm
}

\maketitle
\vspace{-0.2cm} 
\begin{figure}[h]

\makebox[0pt][l]{\hspace{-0.5cm}\includegraphics[width=1.06\textwidth]{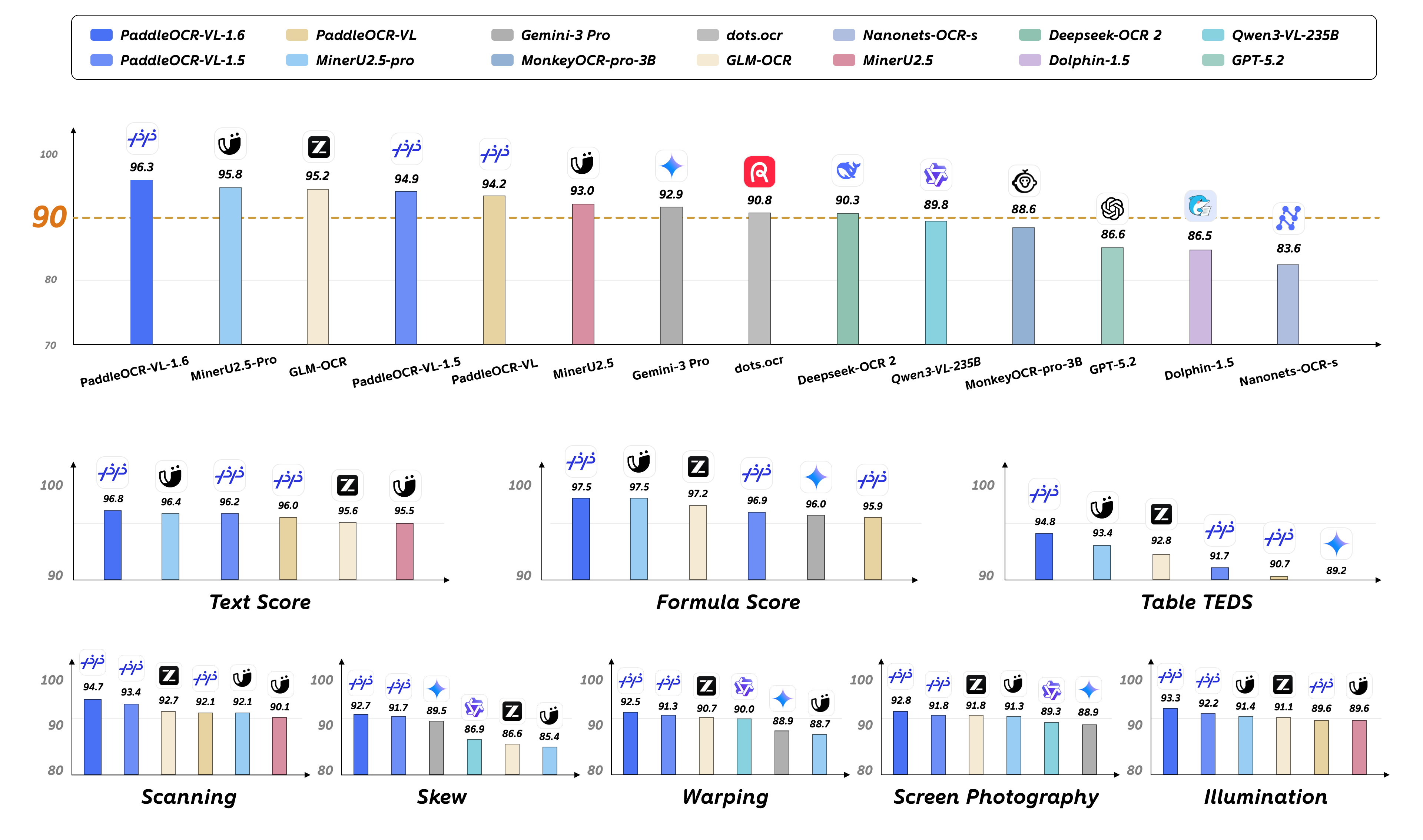}}

\caption{
    Performance of PaddleOCR-VL-1.6 on OmniDocBench v1.6 and Real5-OmniDocBench. 
}
\label{fig:dataset}
\end{figure}

\clearpage
\restoregeometry

\newpage
\setlength{\cftbeforesecskip}{6pt}   
\setlength{\cftbeforesubsecskip}{4pt} 
\setcounter{tocdepth}{2}
\tableofcontents

\newpage

\section{Introduction}

Document parsing has become a central interface between unstructured documents and large language model applications. Modern document systems are expected to recover not only plain text, but also layout regions, reading order, mathematical formulas, tables, charts, seals, and spatially grounded text instances. This structured conversion determines whether document collections can be faithfully transformed into Markdown, JSON, or other machine-readable formats for downstream indexing, retrieval, and reasoning. As retrieval-augmented generation systems increasingly depend on high-fidelity document ingestion~\cite{lewis2020retrieval}, document parsing has moved from a narrow OCR task toward a broader vision-language problem that requires visual localization, structural reconstruction, and semantic preservation across heterogeneous document elements~\cite{li2025monkeyocr,niu2025mineru2,feng2025dolphin,liu2025points}.

Recent progress in document parsing has been driven by both specialized document VLMs and general-purpose multimodal models. PaddleOCR-VL~\cite{cui2025paddleocrvl} showed that a compact 0.9B vision-language model can achieve strong multilingual document parsing performance without relying on larger parameter scales. Other systems, including DeepSeek-OCR~\cite{wei2025deepseek}, MonkeyOCR~\cite{li2025monkeyocr}, Dolphin~\cite{feng2025dolphin}, and HunyuanOCR~\cite{hunyuanvisionteam2025hunyuanocrtechnicalreport}, have further explored end-to-end parsing, heterogeneous prompting, and unified OCR-centric modeling. Building on this line, PaddleOCR-VL-1.5~\cite{cui2026paddleocrvl15multitask09bvlm} strengthened the PaddleOCR-VL series through improved robustness and broader task coverage, while preserving the compact 0.9B model scale. These advances establish a strong starting point for PaddleOCR-VL-1.6: the question is no longer whether compact document parsing VLMs are viable, but how to further improve them once the main architecture has already entered a high-performance regime.

In this regime, the remaining errors are not well described as uniformly distributed noise. Recent benchmark reports and model analyses suggest that top systems increasingly encounter difficult regions that are not fully addressed by increasing data volume or model size alone~\cite{cui2025paddleocrvl,niu2025mineru2,ouyang2025omnidocbench}. Long-tail document layouts, rare scripts, dense formulas, complex tables, and noisy supervision can remain underrepresented or unreliable even when the overall training corpus is large. PaddleOCR-VL-1.5~\cite{cui2026paddleocrvl15multitask09bvlm} already incorporated uncertainty-aware sampling and distortion-oriented robustness improvements, which helped expose the value of targeted data construction. PaddleOCR-VL-1.6 extends this direction by treating the remaining problem as one of under-optimized regions: localized parts of the data and supervision space where the model is unstable, insufficiently covered, or trained against labels that may not be reliable.

To address this problem, we introduce an Under-Optimized Region driven data engine. The engine starts from PaddleOCR-VL-1.5 and diagnoses three complementary types of residual regions. Boundary-fragile regions contain samples whose predictions vary across training checkpoints or under semantic-preserving visual perturbations, indicating unstable decision boundaries. Coverage-sparse regions correspond to low-density neighborhoods in feature semantic space, where long-tail document patterns are likely to be absorbed by dominant distributions under conventional sampling. External-support-deficient regions identify existing training samples whose labels cannot be supported by independent expert parsers, revealing unreliable supervision rather than merely difficult inputs. These signals are then handled through two routes. Boundary-fragile and coverage-sparse samples serve as seeds for region-guided retrieval from internal large document pools, thereby strengthening these underrepresented distributions with minimal disruption to the existing data distribution. External-support-deficient samples are used for existing-label correction. Retrieved unlabeled samples are labeled through expert consensus, and unresolved data are further processed with an Iterable Judge-and-Refine labeling strategy.

The curated data produced by this engine are used in a progressive post-training recipe rather than a single mixed training stage. The Continued Pre-training stage incorporates all curated data to inject broad distributional coverage and corrected supervision into the model. The Supervised Fine-Tuning stage then focuses on high-difficulty and high-quality samples, sharpening model behavior in regions where PaddleOCR-VL-1.5 remains fragile or previously learned from unreliable labels. Finally, GRPO~\cite{shao2024deepseekmath} is applied to further improve model performance. Since data efficiency is critical for reinforcement learning on compact models, we adopt a carefully designed GRPO-oriented data selection strategy. Specifically, candidate samples are jointly assessed from three perspectives: improvement potential, entropy-based uncertainty, and rollout reward distributions. Only high-value samples with the greatest expected gains are selected for the final reinforcement learning stage. 

PaddleOCR-VL-1.6 actively addresses current challenges in document processing with a high-performance, resource-efficient multimodal document parsing solution. Its key contributions include:

\begin{itemize}
    \item We introduce PaddleOCR-VL-1.6, an upgraded version of PaddleOCR-VL-1.5~\cite{cui2026paddleocrvl15multitask09bvlm} built upon improved data strategies and a refined post-training pipeline. It preserves the high-efficiency compact 0.9B model scale while achieving state-of-the-art performance on OmniDocBench v1.6.
    
    \item We introduce Under-Optimized Region Mining, which diagnoses model-specific boundary-fragile, coverage-sparse, and unreliable-supervision regions. We further develop a high-precision automatic annotation pipeline that combines multi-expert consensus with an Iterable Judge-and-Refine labeling strategy, enabling large-scale labeling of unlabeled samples.
    
    \item We design a reliable data selection strategy for reinforcement learning on compact models, where data quality is particularly critical. Candidate samples are evaluated from three complementary perspectives: improvement potential, entropy-based uncertainty, and rollout reward distributions, ensuring effective reinforcement learning for compact models.
    
    \item We develop a progressive CPT-SFT-RL post-training recipe for the PaddleOCR-VL series, providing a practical reference for efficient adaptation to downstream domain-specific scenarios.
\end{itemize}

\section{PaddleOCR-VL-1.6 Overview}
\label{Architecture PaddleOCR-VL}

PaddleOCR-VL-1.6 continues the compact design philosophy of the PaddleOCR-VL series. The full system consists of two models: PP-DocLayoutV3 for layout analysis and PaddleOCR-VL-1.6-0.9B for vision-language understanding. In this upgrade, we keep PP-DocLayoutV3 unchanged and focus on improving the PaddleOCR-VL-1.6-0.9B model. 

PaddleOCR-VL-1.6-0.9B inherits the lightweight architecture of PaddleOCR-VL-1.5-0.9B~\cite{cui2025paddleocrvl}, integrating a Native Resolution Visual Encoder~\cite{dehghani2023patch}, an Adaptive MLP Connector, and the lightweight ERNIE-4.5-0.3B Language Model~\cite{ernie2025technicalreport}. The main upgrade lies not in enlarging the model or modifying the architecture, but in a more targeted data engine and a refined post-training process. This design allows PaddleOCR-VL-1.6 to retain the high inference efficiency of PaddleOCR-VL-1.5 while achieving stronger system performance.

\begin{figure}[H]
\centering
\includegraphics[width=\linewidth]{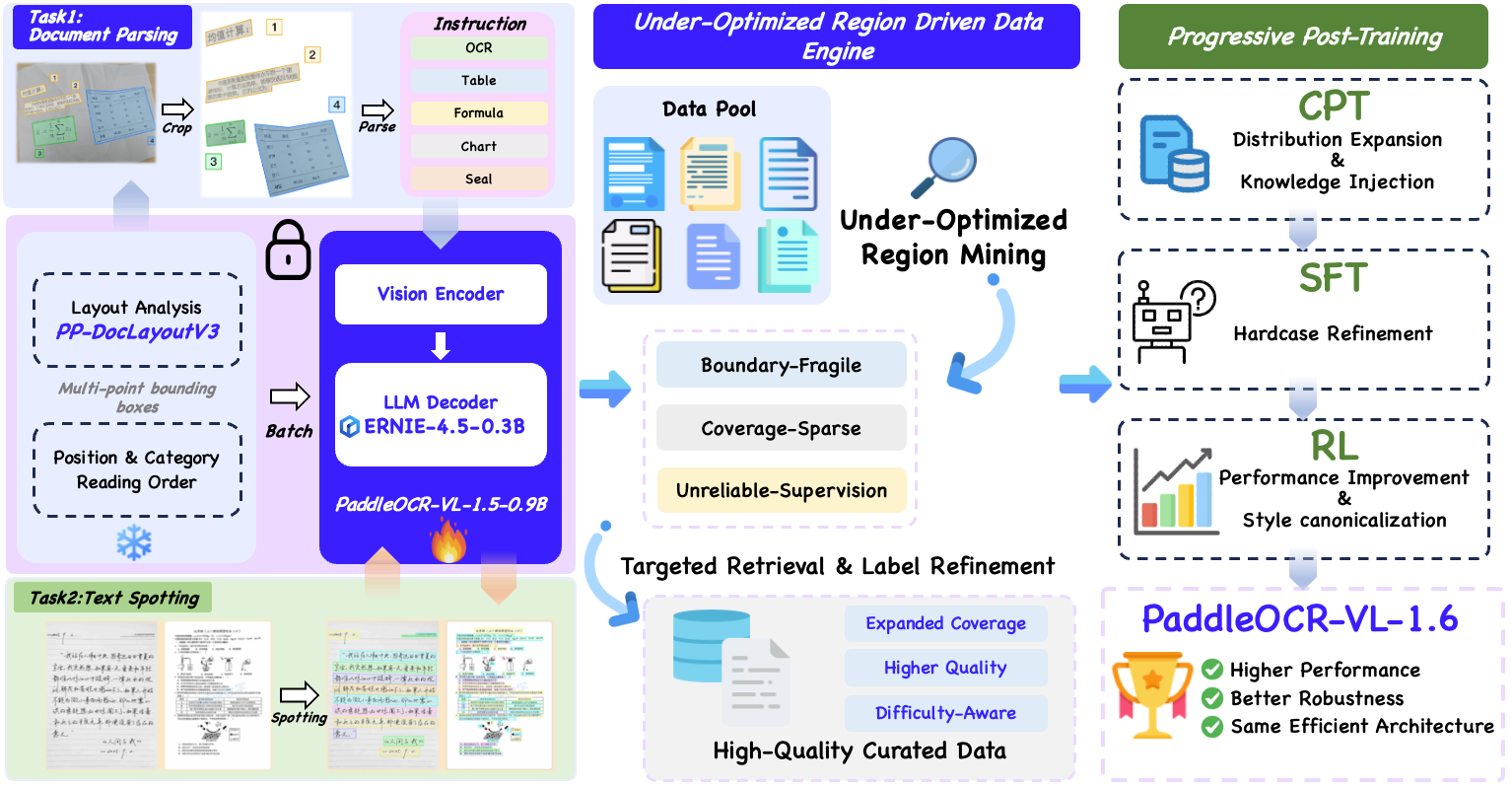} 

\caption{
    \centering
    The overview of PaddleOCR-VL-1.6.
}
\label{fig:model_overview}
\end{figure}

Consistent with its predecessor, PaddleOCR-VL-1.6 supports two primary practical tasks: document parsing and text spotting. For document parsing, the system follows a robust two-stage framework. In the first stage, PP-DocLayoutV3 performs high-precision layout analysis and supports multi-point localization, enabling accurate region localization under complex real-world conditions such as perspective distortion, curved pages, or irregular document layouts. In the second stage, PaddleOCR-VL-1.6-0.9B recognizes the localized regions across diverse document elements, including text, tables, formulas, charts, and seals. A lightweight post-processing engine then organizes these outputs into structured formats such as Markdown and JSON, with additional support for cross-page table merging and heading hierarchy refinement.

For text spotting, PaddleOCR-VL-1.6 directly uses PaddleOCR-VL-1.6-0.9B for end-to-end text detection and recognition. This streamlined workflow supports a broad range of scenarios, including standard documents, identification cards, ancient manuscripts, advertising posters, dialogue screenshots, signboards, and multilingual text images.

The main difference from its predecessor lies in how PaddleOCR-VL-1.6 is improved. PaddleOCR-VL-1.5 expanded robustness and task coverage, whereas PaddleOCR-VL-1.6 focuses on the residual weaknesses that remain after this strong baseline. Its development process starts by diagnosing under-optimized regions from PaddleOCR-VL-1.5, including samples with fragile predictions, sparse distributional coverage, and unreliable existing labels. These diagnostic signals guide data construction and refinement rather than being treated as isolated evaluation failures. As shown in Figure~\ref{fig:model_overview}, the upgrade path of PaddleOCR-VL-1.6 is organized around data engineering and post-training: identifying residual weak regions, applying targeted enhancements to improve model performance in these regions, and using a staged optimization recipe that matches each data subset's reliability and learning value.

At a high level, PaddleOCR-VL-1.6 contains three key components. The first component is an under-optimized-region-driven data engine. It discovers boundary-fragile and coverage-sparse regions as retrieval seeds for new unlabeled samples, while using external-support-deficient regions to detect unreliable annotations in the existing training set. The second component is expert-consensus labeling and refinement. Retrieved samples are labeled by multiple expert parsers, and hard cases for which expert consensus remains insufficient are further refined through an iterative Judge-and-Refine process. The third component is progressive post-training, which follows a complete CPT-SFT-RL pipeline and serves as a practical training recipe for the PaddleOCR-VL series. Before the RL stage, we also develop a standardized and reusable selection strategy to identify high-value samples for reinforcement learning. These components will be detailed in the following sections.

\section{Under-Optimized Region Driven Data Engine}

\subsection{Motivation: From Uniform Scaling to Under-Optimized Region Optimization}

PaddleOCR-VL-1.6 starts from a setting in which its predecessor is already a strong baseline. PaddleOCR-VL-1.5 preserves the compact 0.9B scale of the PaddleOCR-VL series while extending robustness and task coverage. In such a high-performance regime, the remaining errors are not well explained by a simple shortage of generic document data. Uniformly enlarging the training corpus may still introduce useful variation, but it also spends limited training budget on regions where the model already behaves reliably. This issue is especially important for compact models such as the PaddleOCR-VL series, whose final performance is more sensitive to data efficiency and distributional balance. Compared with uniform data scaling, targeted data expansion is therefore a more effective strategy in both training efficiency and final model performance.

\begin{figure}[H]
\centering
\includegraphics[width=\linewidth]{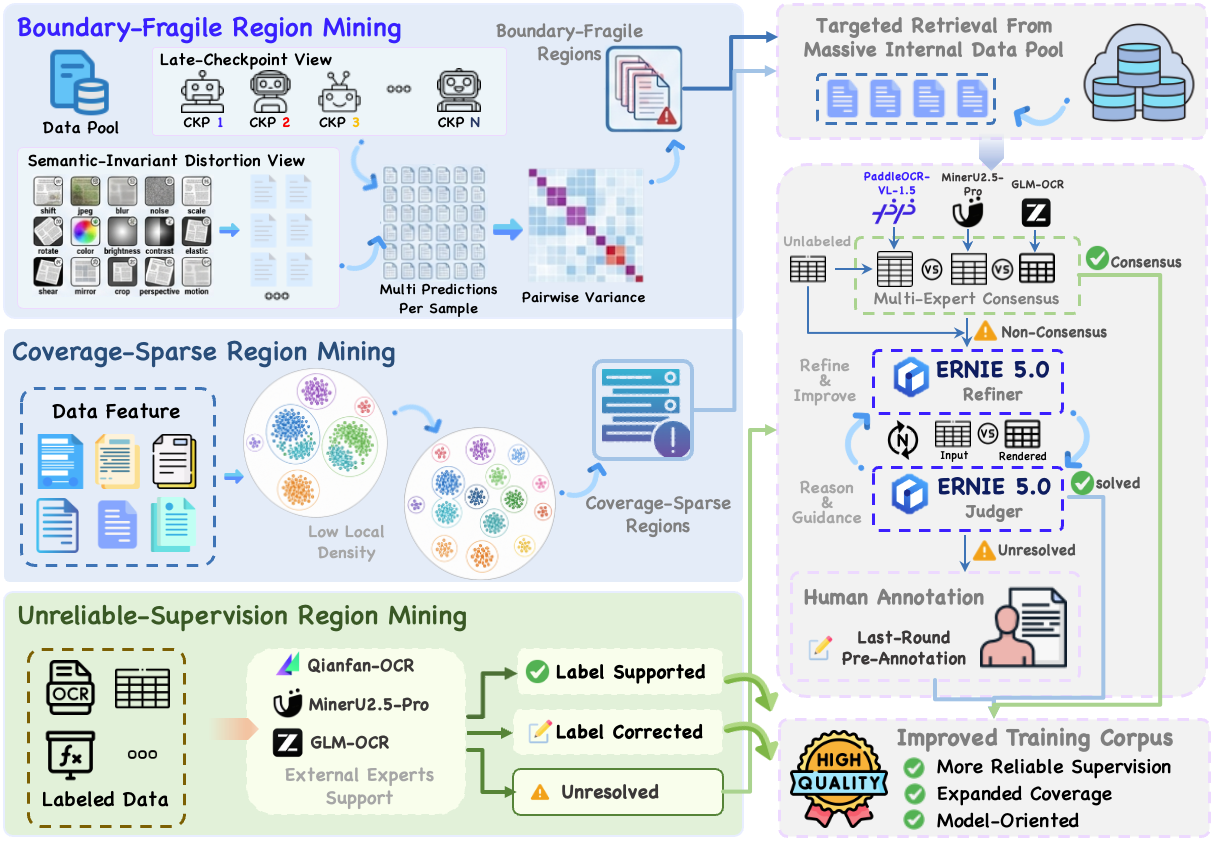} 

\caption{
    \centering
    The overview of PaddleOCR-VL-1.6 Data Engine.
}
\label{fig:engine_overview}
\end{figure}

Our analysis of PaddleOCR-VL-1.5 reveals three characteristic failure patterns. First, small pixel-level shifts or semantic-preserving visual distortions can cause large changes in model outputs, and in some cases even lead to severe degradation. Such failures are difficult to eliminate by simply adding standard data augmentation during training, suggesting that the model has not learned a stable mapping in the corresponding local data region. Second, some samples that already appear in the training distribution are still predicted incorrectly, indicating that the surrounding distribution is insufficiently represented and remains under-optimized. Third, the model sometimes produces stable but incorrect outputs with high confidence, which suggests that the issue lies not only in difficult inputs but also in unreliable supervision signals that have biased the learned mapping.

These observations motivate a model-oriented view of data optimization. Rather than treating all additional data as equally valuable, PaddleOCR-VL-1.6 focuses on Under-Optimized Regions (UORs), namely regions in the data and supervision space where the current model has not yet obtained a reliable mapping from document images to structured outputs. We identify three types of UORs corresponding to the failure patterns above: Boundary-Fragile Regions, where predictions are unstable under small semantic-preserving distortions; Coverage-Sparse Regions, where the local distribution is insufficiently covered by existing data; and Unreliable-Supervision Regions, where the model has learned from unreliable supervision signals. Based on these observations, we build an Under-Optimized-Region-driven data engine to explicitly mine and refine the current model's weak regions, enabling targeted data optimization for PaddleOCR-VL-1.6, as depicted in Figure~\ref{fig:engine_overview}.

\subsection{Boundary-Fragile Regions}

Boundary-Fragile Regions refer to samples for which the model has not formed a stable mapping from document images to structured outputs. These regions are harmful because they make the final converged model less reliable: even under similar training settings, small changes in the optimization trajectory may lead to noticeably different predictions, and the model may behave unstably in certain scenarios. A common way to improve such robustness is to introduce data augmentation and encourage consistency under input variations. However, in our experiments, even a combination of more than ten augmentation operations could not fully eliminate the instability for some samples. This suggests that the issue is not merely a lack of generic augmentation, but that the model remains intrinsically unstable on the local distribution represented by these samples. Therefore, we need a flexible strategy to identify the unstable regions of a given model under its own architecture and data distribution.

We propose Boundary-Fragile Region mining as a model-oriented strategy for locating such unstable regions. The method is designed to be general: for different combinations of model architectures and training data distributions, it can identify regions where the current model has not yet learned robust invariance. Specifically, we evaluate boundary fragility from two complementary views. The first view examines prediction variation across late-stage training checkpoints, where the overall model performance has mostly converged. The second view examines prediction variation under semantic-invariant input distortions applied to the same checkpoint. Together, these two views capture instability induced by both model-state variation and input-appearance variation.

\textbf{View 1: Checkpoint-Level Instability.} The checkpoint view is based on the observation that, near the end of training, the learning rate has annealed to a low level and the global performance of the model has largely stabilized. For well-learned regions, predictions from nearby late-stage checkpoints should therefore remain consistent. In Boundary-Fragile Regions, however, the model can still change its output substantially even when the checkpoint difference is small. Based on this intuition, we retain eight checkpoints from the last 8\% of the training schedule and use their prediction discrepancies to measure checkpoint-level boundary fragility.

\textbf{View 2: Semantic-Invariant Perturbation Sensitivity.} The semantic-invariant distortion view directly measures whether the model is robust to small visual changes that should not alter document semantics. For each checkpoint, we apply a set of mild perturbations to the same input and compare the resulting predictions. These perturbations include pixel shifts, JPEG compression, noise, blur, non-uniform scaling, and other lightweight transformations, forming 16 semantic-invariant distortion types in total. If the structured output changes substantially under these distortions, the sample indicates a local region where the model has not learned stable invariance.

For PaddleOCR-VL-1.5-0.9B, we apply this mining strategy to the full training dataset. Each sample is evaluated under the Cartesian product of 8 late-stage checkpoints and 16 semantic-invariant distortions, resulting in 128 predictions per sample. We then serialize the predictions into task outputs and compute the normalized edit distance for every prediction pair, yielding \((128 \times 127) / 2 = 8128\) pairwise discrepancy scores. To focus on the most significant variations while reducing the effect of minor formatting differences, we select the largest 128 pairwise distances and average them as the Boundary-Fragility Score. In the final selection, we empirically select the top 1\% samples ranked by this score, and additionally include samples for which any of the 128 predictions exhibits model degeneration.

Through this two-view mining process, PaddleOCR-VL-1.6 identifies Boundary-Fragile Regions from both late-checkpoint instability and semantic-invariant perturbation sensitivity. These samples reveal local distributions where the current model remains unreliable, and they serve as targeted anchors for subsequent data retrieval and refinement.

\subsection{Coverage-Sparse Regions}

Coverage-Sparse Regions address a different failure mode. As discussed above, some samples may still be predicted incorrectly even when similar patterns have appeared in the training corpus. These failures are not necessarily caused by an unstable decision boundary; instead, they often arise because the surrounding distribution is weakly supported by existing data. Uniform data scaling may introduce more samples overall, but without a mechanism for recognizing sparse neighborhoods, it can continue to over-sample dominant distributions while leaving long-tail regions underrepresented. Therefore, PaddleOCR-VL-1.6 requires an explicit strategy to locate Coverage-Sparse Regions in the current training distribution.

PaddleOCR-VL-1.6 diagnoses coverage sparsity through a visual-semantic neighborhood view. The data engine first extracts representations for all training samples using an internal document-specific feature encoder. It then measures sample similarity in the resulting feature space and discovers small, weakly connected outlier clusters as candidate Coverage-Sparse Regions. These clusters indicate local neighborhoods where the current corpus provides insufficient distributional support.

\begin{algorithm}[H]
\caption{Coverage-Sparse Region Mining}
\label{alg:coverage_sparse_mining}
\begin{algorithmic}[1]
\REQUIRE Training samples $\mathcal{D}=\{x_i\}_{i=1}^{N}$, document-specific feature encoder $f(\cdot)$, target number of clusters $K_{target}$, initial similarity threshold $\tau_0$, threshold step size $\Delta\tau$.
\ENSURE Candidate Coverage-Sparse Regions $\mathcal{R}_{cs}$.
\STATE Extract normalized document features $z_i = f(x_i)$ for all $x_i \in \mathcal{D}$.
\STATE Compute pairwise cosine similarities $s_{ij}=z_i^{\top}z_j$.
\STATE Build an initial similarity graph $G=(V,E)$, where $V=\mathcal{D}$ and $E=\{(i,j)\mid s_{ij} \geq \tau_0\}$.
\STATE Obtain connected components $\mathcal{C}$ from $G$ and set $\tau \leftarrow \tau_0$.
\WHILE{$|\mathcal{C}| < K_{target}$}
    \STATE Update the threshold $\tau \leftarrow \tau + \Delta\tau$.
    \STATE Initialize $\mathcal{C}_{new} \leftarrow \emptyset$.
    \FOR{each component $C \in \mathcal{C}$}
        \STATE Build $G_C=(C,E_C)$, where $E_C=\{(i,j)\mid x_i,x_j\in C,\ s_{ij}\geq\tau\}$.
        \STATE Split $C$ into connected components of $G_C$ and add them to $\mathcal{C}_{new}$.
    \ENDFOR
    \STATE Update $\mathcal{C} \leftarrow \mathcal{C}_{new}$.
\ENDWHILE
\STATE Select small outlier components from $\mathcal{C}$ as $\mathcal{R}_{cs}$.
\RETURN $\mathcal{R}_{cs}$.
\end{algorithmic}
\end{algorithm}

As shown in Algorithm~\ref{alg:coverage_sparse_mining}, the method gradually increases the similarity threshold to reveal fine-grained clusters. Instead of forcing all samples into a fixed partition at once, it progressively splits the similarity graph and identifies small outlier components that has low local density.

This density-oriented clustering strategy is well suited to Coverage-Sparse Region mining. The goal is not to obtain balanced semantic clusters, but to expose weakly supported tail neighborhoods that are easily hidden by dominant distributions. In contrast, fixed-$K$ clustering methods such as K-Means require the number of clusters to be specified in advance and assign every sample to a cluster, which can cause rare document modes to be absorbed into nearby dense groups. By preserving neighborhood connectivity, our method keeps sparse regions visible and uses them as targeted data expansion seeds. Based on the mined Coverage-Sparse Regions, PaddleOCR-VL-1.6 systematically supplements long-tail data such as ancient books, rare characters, and industrial tables, further improving the model's capability on underrepresented scenarios.

\subsection{Unreliable-Supervision Regions}

The previous two mining strategies mainly identify weak regions that require distributional expansion: Boundary-Fragile Regions expose locally unstable samples, while Coverage-Sparse Regions reveal underrepresented neighborhoods in the current corpus. In practice, we observe that the model may repeatedly produce the same high-confidence error patterns, suggesting that some failures originate from unreliable supervision rather than insufficient coverage. Therefore, Unreliable-Supervision Region mining focuses on the existing labels themselves, aiming to identify inaccurate targets and improve the overall effectiveness of supervision in the training set.

To diagnose such regions, we introduce an external-support-based verification strategy. The key idea is that high-performance models trained from different data sources and model architectures can provide independent expert views for the same sample, helping break the bias of the original annotation system. Specifically, we use Qianfan-OCR~\cite{QianfanOCR}, GLM-OCR~\cite{duan2026glmocrtechnicalreport}, and MinerU2.5-Pro~\cite{MinerU25Pro} as three expert models, each producing an independent prediction for the same training sample. The original label is then verified or corrected according to expert agreement. If at least one expert prediction agrees with the original label, the sample is regarded as externally supported and the original label is retained. If the original label disagrees with all experts but at least two experts agree with each other, the sample is treated as label-correctable, and the consensus expert output is used to replace the original label. If neither the original label nor the three experts reach agreement, the sample is marked as unresolved and sent to the subsequent fine-grained automatic annotation strategy described in the next section.

This strategy provides a conservative but effective way to repair supervision noise. Through this process, PaddleOCR-VL-1.6 mines and improves Unreliable-Supervision Regions inherited from PaddleOCR-VL-1.5. In addition, the agreement pattern among experts naturally stratifies data difficulty: samples resolved by expert consensus can be used as high-confidence corrected data, while samples without expert agreement are treated as difficult cases and handled carefully in later post-training stages.

\subsection{Automatic Annotation via Multi-Expert Consensus and Render-Guided Refinement}

After Under-Optimized Region mining, the data engine obtains two types of samples that require reliable supervision. Samples retrieved from internal document pools using Boundary-Fragile and Coverage-Sparse seeds are unlabeled. In parallel, samples identified from Unreliable-Supervision Regions may already have labels, but these labels lack sufficient external support and therefore require correction or refinement. PaddleOCR-VL-1.6 introduces a high-precision automatic annotation pipeline that combines multi-expert consensus with render-guided iterative refinement.

For difficult document parsing tasks such as table recognition and formula recognition, label generation often requires stronger test-time reasoning, especially when multiple expert models produce inconsistent outputs. Therefore, as the judge-and-refine model for hard cases, we use ERNIE 5.0~\cite{wang2026ernie}, a natively autoregressive foundation model designed for unified multimodal understanding and generation across text, image, video, and audio, with strong visual reasoning capability. As shown in Algorithm~\ref{alg:judge_refine_annotation}, the pipeline first collects predictions from three expert models, PaddleOCR-VL-1.5~\cite{cui2026paddleocrvl15multitask09bvlm}, GLM-OCR~\cite{duan2026glmocrtechnicalreport}, and MinerU2.5-Pro~\cite{MinerU25Pro}. If at least two experts agree, their consensus output is directly accepted as the label. Otherwise, the sample is treated as difficult and enters the render-guided Judge-and-Refine stage.

The design has two practical details. First, the three expert predictions are injected only during the initial ERNIE 5.0~\cite{wang2026ernie} prediction. Subsequent refinement rounds use only the current prediction and the discrepancies identified by the previous judge step, which prevents repeated expert outputs from biasing the refinement trajectory. Second, the judge step is render-guided rather than purely text-based. For formulas and tables, directly comparing an image with LaTeX or HTML is difficult even for strong multimodal models. Rendering the candidate output converts the comparison into a same-modality visual matching problem, allowing the judge to more accurately locate row or column misalignment, wrong spanning cells, and content placement errors.

\begin{algorithm}[H]
\caption{Multi-Expert Consensus and Render-Guided Label Refinement}
\label{alg:judge_refine_annotation}
\begin{algorithmic}[1]
\REQUIRE Input image $x$, expert models $\{E_1,E_2,E_3\}$, judge-and-refine model $M$, maximum refinement rounds $T$.
\ENSURE Accepted label $y$ or manual-annotation request with pre-label $\hat{y}$.
\STATE Generate expert predictions $\{y_1,y_2,y_3\}$ using $\{E_1,E_2,E_3\}$.
\IF{at least two expert predictions agree}
    \STATE Set $y$ to the consensus expert output.
    \RETURN accepted label $y$.
\ENDIF
\STATE $\hat{y}^{(0)} \leftarrow M_{\mathrm{refine}}(x, y_1, y_2, y_3)$ \COMMENT{initial prediction with expert references}
\FOR{$t = 0$ to $T-1$}
    \STATE Render $\hat{y}^{(t)}$ into an image $r^{(t)}$.
    \STATE $\delta^{(t)} \leftarrow M_{\mathrm{judge}}(x, r^{(t)})$ \COMMENT{detect discrepancies between the input image and rendered prediction}
    \IF{$\delta^{(t)} = \emptyset$}
        \STATE $y \leftarrow \hat{y}^{(t)}$
        \RETURN accepted label $y$
    \ENDIF
    \STATE $\hat{y}^{(t+1)} \leftarrow M_{\mathrm{refine}}(\hat{y}^{(t)}, \delta^{(t)})$ \COMMENT{refine the prediction using detected discrepancies}
\ENDFOR
\RETURN manual-annotation request with the last prediction $\hat{y}^{(T)}$ as the pre-annotation.
\end{algorithmic}
\end{algorithm}

This pipeline enables PaddleOCR-VL-1.6 to automatically produce reliable labels for most difficult samples. Unresolved cases are forwarded to manual annotation, with the final pipeline output used as a pre-annotation to reduce human effort.

\section{Progressive Post-Training Recipe}

PaddleOCR-VL-1.6 is not trained from scratch, it starts from the PaddleOCR-VL-1.5 checkpoint and improves the model through a curated progressive post-training recipe. After the base architecture has reached a strong performance regime, the key objective is to absorb newly constructed high-value data efficiently rather than restart large-scale pre-training.This section describes how data produced by the Under-Optimized Region Driven Data Engine are allocated across three stages. Continued Pre-Training (CPT) absorbs broad curated data to expand distributional coverage and incorporate corrected supervision. Supervised Fine-Tuning (SFT) focuses on high-quality hard samples to refine document parsing behavior. Reinforcement Learning (RL) further optimizes high-potential samples with verifiable rewards. This staged design uses each data subset according to its reliability and learning value, improving PaddleOCR-VL-1.6 while preserving the compact 0.9B model scale.

\subsection{Continued Pre-Training for Distributional Expansion}
\label{training_strategy_cpt}

The first stage is designed to absorb the newly introduced expanded data distribution. Beyond improving the reliability of existing annotations, the data engine brings in a large number of newly retrieved samples from previously under-optimized regions, including ancient books, rare characters, and other long-tail document scenarios. These samples introduce distributional shifts that cannot be fully learned through a narrow supervised fine-tuning stage. Continued Pre-Training (CPT) is therefore used to inject and stabilize these new document patterns before more selective optimization stages. 

\textbf{Training data.} The CPT corpus combines the full SFT data and part of the pre-training data from PaddleOCR-VL-1.5 with all newly retrieved data produced by the data engine, resulting in 16.8M training samples. All samples use the latest annotations, providing both broader coverage and higher-quality supervision.

\textbf{Training settings.} All model parameters are unfrozen to adapt to the expanded distribution. We train for one epoch with a global batch size of 1024 and set the maximum learning rate to $3\times10^{-5}$ for all parameters.

\subsection{Supervised Fine-Tuning for Hard-Case Refinement}
\label{training_strategy_sft}
CPT expands the model's distributional coverage and establishes its foundational capabilities, SFT further refines the model on difficult samples with reliable supervision. The goal of this stage is not to reuse all curated data uniformly, but to concentrate supervised learning on cases where the model still requires stronger task behavior.

\textbf{Training data.} The SFT corpus was constructed from three sources. First, we follow the Uncertainty-Aware Cluster Sampling (UACS) strategy used in PaddleOCR-VL-1.5~\cite{cui2026paddleocr} to mine hard samples from the CPT corpus. Second, we include all samples for which the three experts fail to reach agreement and therefore enter the Render-Guided Refinement pipeline. These samples are difficult by construction and require further supervised learning after reliable labels are obtained. Third, we include samples originally present in the PaddleOCR-VL-1.5 training data whose labels are identified and corrected through Unreliable-Supervision Region mining. In total, this process selects 7.3M samples for SFT.

\textbf{Training settings.} All model parameters are unfrozen. We train for one epoch with a global batch size of 1024 and set the maximum learning rate to $1\times10^{-5}$ for all parameters. 

\subsection{Reinforcement Learning for High-Potential Optimization}
\label{training_strategy_rl}

Reinforcement learning (RL) provides an additional optimization signal for beyond supervised learning. The training corpus contains large-scale data from different sources and annotation styles, the model may produce multiple output styles for similar input patterns. RL helps regularize these behaviors. It also further improves model performance and generalization while suppressing degeneration on out-of-distribution samples.

Applying RL to PaddleOCR-VL-1.6-0.9B, however, requires careful data selection. The language model component is only 0.3B, making the compact model more sensitive to RL data quality and sample efficiency. If RL samples are selected with a casual strategy, the model may improve on a subset of hard cases while degrading overall performance. Therefore, the RL stage must focus on samples that are both learnable and likely to produce meaningful reward-driven gains. To address this, we propose a GRPO-oriented High-Potential Sample Mining strategy for selecting effective RL training samples, thereby stabilizing the RL training process and ensuring the effectiveness of reward-driven optimization.

\subsubsection{GRPO-oriented High-Potential Sample Mining}

GRPO~\cite{shao2024deepseekmath} optimizes the policy by comparing multiple sampled responses for the same input and assigning advantages according to their relative rewards within the group. This group-relative formulation removes the need for a separate value model, but it also makes the effectiveness of training highly dependent on whether each prompt can produce informative reward differences. For PaddleOCR-VL-1.6-0.9B, this requirement is particularly important because the language model component is compact, making the policy more sensitive to noisy, over-easy, over-hard, or reward-flat RL samples.

We therefore introduce a GRPO-oriented high-potential sample mining strategy to select RL data according to the current SFT policy. The SFT model is used as the initial policy to probe a candidate RL data pool. For each candidate sample \(x\), we generate 16 rollouts with temperature \(0.85\), top-\(p=0.9\), and top-\(k=32\). Each rollout is evaluated by the task-specific verifiable reward function described in the following section, producing an empirical reward distribution for the sample.

\textbf{Non-informative sample filtering.} The first step is to remove samples that are unlikely to contribute useful GRPO updates. Overly difficult samples are filtered when the maximum rollout reward \(r_{\max}(x)\) is below a threshold, since the current policy never reaches a sufficiently good output and the reward signal mainly indicates failure. Overly easy samples are filtered when the mean reward \(r_{\mathrm{mean}}(x)\) is above a threshold, since the model has already solved them with little remaining headroom. We further define the learning potential of a sample as \(r_{\max}(x)-r_{\mathrm{mean}}(x)\). A small gap indicates that even the best sampled output is not meaningfully better than the average rollout, so the sample provides limited opportunity for reward-driven improvement. Finally, samples with very low reward variance are removed because GRPO relies on relative reward differences within the sampled group; reward-flat rollouts provide weak or degenerated advantage signals.

\textbf{High-potential sample scoring.} For the remaining candidates, we compute a unified high-potential score that combines improvement headroom, generation uncertainty, and reward diversity. The dominant term is the learning-potential gap \(r_{\max}(x)-r_{\mathrm{mean}}(x)\), which measures whether the current policy can occasionally produce outputs substantially better than its average behavior. We also estimate generation uncertainty from the likelihood of sampled rollouts under the current policy. For the \(k\)-th rollout \(y^{(k)}=(y^{(k)}_1,\ldots,y^{(k)}_{T_k})\), we define its length-normalized sequence confidence as
\begin{equation}
C\left(y^{(k)} \mid x\right)
=
\left(
\prod_{t=1}^{T_k}
p_{\theta}\left(y^{(k)}_t \mid x, y^{(k)}_{<t}\right)
\right)^{1/T_k}.
\end{equation}
This geometric mean removes the length bias of raw sequence likelihood and measures how confidently the current policy generates the rollout at the token level. The sample-level uncertainty is then computed by averaging over the \(K\) rollouts:
\begin{equation}
U(x)
=
1 -
\frac{1}{K}
\sum_{k=1}^{K}
C\left(y^{(k)} \mid x\right),
\quad K=16.
\end{equation}
A larger \(U(x)\) indicates that the current policy assigns lower average confidence to its sampled outputs on \(x\), suggesting that the generation behavior is not yet stable and may still benefit from policy refinement.

In addition, we use the reward variance to measure whether the sampled rollouts expose meaningful distinctions under the task reward:
\begin{equation}
V_r(x)
=
\frac{1}{K}
\sum_{k=1}^{K}
\left(
r^{(k)}(x) - r_{\mathrm{mean}}(x)
\right)^2,
\end{equation}
where \(r^{(k)}(x)\) is the reward of the \(k\)-th rollout and \(r_{\mathrm{mean}}(x)=\frac{1}{K}\sum_{k=1}^{K}r^{(k)}(x)\). While \(U(x)\) captures uncertainty in the generation process, \(V_r(x)\) captures diversity in task-level outcomes, which is directly relevant to group-relative optimization.

The final high-potential score is defined as
\begin{equation}
\mathrm{Score}(x)
=
\left(r_{\max}(x)-r_{\mathrm{mean}}(x)\right)
\cdot
\exp\left(\alpha U(x)+\beta V_r(x)\right),
\end{equation}
where \(r_{\max}(x)=\max_{k} r^{(k)}(x)\), and \(\alpha\) and \(\beta\) control the contributions of generation uncertainty and reward variance (we set \(\alpha = 1\) and \(\beta = 2\) in practice), respectively. The leading term \(r_{\max}(x)-r_{\mathrm{mean}}(x)\) measures the reachable improvement headroom of the sample, while the exponential factor upweights samples whose rollouts are both uncertain under the current policy and discriminative under the task reward.This formulation prioritizes samples that are not merely difficult, but learnable: the policy can already reach a better solution in some rollouts, the reward distribution provides discriminative group-relative signals, and the generation process still has sufficient uncertainty to benefit from optimization.

To preserve task balance, this scoring and selection process is performed separately for all tasks, including OCR, chart parsing, table recognition, formula recognition, seal recognition, and text spotting. The top-ranked samples from each task are then used for the final GRPO stage. In this way, RL training focuses on high-quality candidates with observable improvement potential, rather than uniformly sampling from the entire candidate pool. This stabilizes GRPO optimization and makes reward-driven learning more effective for the compact PaddleOCR-VL-1.6-0.9B model.

\subsubsection{Reward Design}

For a compact model such as PaddleOCR-VL-1.6-0.9B, overly sparse binary rewards provide limited learning signals, making it difficult for the model to benefit effectively from RL. We therefore design a representation-aware verifiable reward that provides task-aligned scalar feedback while still enforcing strict validity constraints. For each task $t$, the model output $y$ and reference $y^*$ are first mapped into a task-specific canonical representation by $\phi_t$. The final reward is defined as

\begin{equation}
R_t(y, y^*) =
\mathrm{Valid}_t(y)
\cdot
\mathrm{Struct}_t(\phi_t(y))
\cdot
\mathrm{Sim}_t(\phi_t(y), \phi_t(y^*)),
\label{eq:representation_aware_reward}
\end{equation}

where $\mathrm{Valid}_t$ is a strict validity gate, $\mathrm{Struct}_t$ is a structural adjustment factor, and $\mathrm{Sim}_t$ is the task-aligned similarity metric. The validity gate defines the minimum requirement for a usable task output and is binary: outputs with invalid format, malformed LaTeX, truncation, degeneration, or other task-specific failures receive zero reward. The structural factor softly penalizes outputs that are parsable but require post-processing correction. For example, non-rectangular OTSL table outputs are penalized according to the minimum edit cost required to convert them into valid rectangular structures. The similarity term then measures how close the valid, canonicalized output is to the reference using the metric appropriate for each task. The task-specific reward designs are summarized in Table~\ref{tab:representation_aware_reward}.

\begin{table}[t]
\centering
\small
\setcellgapes{2pt}
\makegapedcells
\renewcommand{\arraystretch}{1.15}
\begin{tabular}{p{0.13\linewidth} p{0.30\linewidth} p{0.25\linewidth} p{0.20\linewidth}}
\toprule
Task & $\mathrm{Valid}_t$ & $\mathrm{Struct}_t$ & $\mathrm{Sim}_t$ \\
\midrule
Table &
\makecell[l]{Model Degeneration \\[1pt] Output Truncation \\[1pt] Unparsable OTSL Output} &
\makecell[l]{Cell-Level LaTeX Validity \\[1pt] OTSL Rectangularity Cost} &
TEDS \\[0.6em]
OCR &
\makecell[l]{Model Degeneration \\[1pt] Output Truncation \\[1pt] Invalid Inline LaTeX} &
$\mathrm{Struct}_t = 1$ &
$1-\mathrm{NED}$ \\[0.6em]
Formula &
\makecell[l]{Model Degeneration \\[1pt] Output Truncation \\[1pt] Invalid LaTeX Syntax} &
$\mathrm{Struct}_t = 1$ &
CDM \\[0.6em]
Seal &
\makecell[l]{Model Degeneration \\[1pt] Output Truncation} &
$\mathrm{Struct}_t = 1$ &
$1-\mathrm{NED}$ \\[0.6em]
Chart &
\makecell[l]{Model Degeneration \\[1pt] Output Truncation \\[1pt] Invalid Markdown Table \\[1pt] Malformed Rows or Columns} &
Table Rectangularity Cost &
RMS-F1 \\[0.6em]
Spotting &
\makecell[l]{Model Degeneration \\[1pt] Output Truncation \\[1pt] Invalid Output Format} &
$\mathrm{Struct}_t = 1$ &
Edit-Similarity-Weighted F1 Score \\
\bottomrule
\end{tabular}
\caption{Reward design for PaddleOCR-VL-1.6. Each task follows the same Valid-Struct-Sim formulation while using task-specific validity checks, structural factors, and similarity metrics.}
\label{tab:representation_aware_reward}
\end{table}

Specifically, for text spotting, each geometrically matched prediction-reference box pair is weighted by text similarity, using $1-\mathrm{NED}$ between the predicted and reference strings. This yields an edit-similarity-weighted F1 score that jointly rewards accurate localization and recognition, instead of treating all matched boxes as equally correct.

\subsubsection{Training Data and Settings}

\noindent\hspace*{\parindent}\textbf{Training data.} We build a carefully curated RL candidate data pool with unified annotation styles, high-quality references, and challenging samples that can provide meaningful reward signals. Using the SFT model as the rollout policy, we apply the high-potential sample mining strategy described above to probe, filter, and score samples in this candidate pool. For each task, we empirically select the top 8K samples according to the final mining score for GRPO training. The resulting RL training set contains 49K samples in total.

\noindent\hspace*{\parindent}\textbf{Training settings.} All model parameters are unfrozen during the RL stage. We train for two epochs with a global batch size of 1024 and set the maximum learning rate to $2\times10^{-6}$ for all parameters. During rollout sampling, we use a temperature of 0.85, top-$k$ of 32, top-$p$ of 0.9, and a group size $G$ of 16. Following DAPO~\cite{yu2026dapo}, we adopt a clip-higher strategy with $\epsilon_{\mathrm{high}}=0.28$. We also use the dynamic sampling strategy from DAPO to ignore groups whose within-group reward variance is zero, ensuring that GRPO updates are computed only from samples with meaningful relative reward differences.

\section{Evaluation}

\label{sec:experiments} 

To thoroughly assess the effectiveness of PaddleOCR-VL-1.6, we conducted evaluations on the document parsing benchmark OmniDocBench v1.6\cite{MinerU25Pro} and Real5-OmniDocBench\cite{zhou2026real5omnidocbenchfullscalephysicalreconstruction}. Furthermore, we expanded the evaluation scope by incorporating hard table recognition, chart parsing, text spotting and seal recognition tasks to provide a more comprehensive analysis of the model's performance in practical and complex scenarios.

\subsection{Document Parsing}
\label{Document Parsing}

This section details the evaluation of end-to-end document parsing capabilities using the following two benchmarks, aiming to measure its overall performance in real-world document scenarios. 

\paragraph{OmniDocBench v1.6} We also evaluate on OmniDocBench v1.6, an updated version of OmniDocBench v1.5. Compared with v1.5, v1.6 introduces two key changes. First, it adopts Multi-Granularity Adaptive Matching (MGAM) to reduce matching bias caused by fixed-granularity one-to-one element matching. This improves the robustness of evaluation when a prediction uses a different but semantically equivalent segmentation from the ground truth. Second, it adds a dedicated Hard subset of 296 pages covering more challenging document parsing scenarios, including complex nested tables, dense formula layouts, and unconventional document structures. OmniDocBench v1.6 therefore provides a more comprehensive evaluation. The evaluation metrics remain task-specific. Text and reading order are evaluated using edit-distance-based similarity, tables are evaluated using TEDS, and formulas are evaluated using CDM~\cite{Wang_cdm_2025_CVPR}. With MGAM, these metrics are computed under an adaptive matching strategy that mitigates segmentation-granularity mismatch, and the final score is aggregated over the evaluated document elements.

Table~\ref{tab:omni16_full_performance} demonstrates that PaddleOCR-VL-1.6 achieves state-of-the-art overall performance, consistently outperforming both existing general-purpose VLMs and specialized document parsing models. Notably, PaddleOCR-VL-1.6 delivers a substantial performance leap over its predecessor, PaddleOCR-VL-1.5, raising the overall score from 94.93\% to a top-ranking 96.33\%. Specifically, it achieves improvements of 0.5\%, 0.6\%, 3.09\%, and 2.74\% in Text-Edit distance, CDM Score, Table-TEDS, and Table-TEDS-Structure, respectively. Furthermore, our model establishes new state-of-the-art results in major parsing sub-tasks, including a reduced Text-Edit distance of 0.033, an improved Formula-CDM score of 97.49\%, and leading scores of 94.76\% and 97.11\% in Table-TEDS and Table-TEDS-S, respectively. It also achieves a highly competitive Reading Order score of 0.127, which is comparable to the best-performing models on this metric. These improvements underscore the model's enhanced precision in text recognition, formula extraction, and complex table structure analysis.

\begin{table}[!t]
    \centering

    \resizebox{\textwidth}{!}{%
    \renewcommand{\arraystretch}{1.2}
    \begin{tabular}{l|ll|c|c c c c c}
        \toprule
        \textbf{Model Type} & \textbf{Methods} & \textbf{Parameters} & \textbf{Overall$\uparrow$} & \textbf{Text\textsuperscript{Edit}$\downarrow$} & \textbf{Formula\textsuperscript{CDM}$\uparrow$} & \textbf{Table\textsuperscript{TEDS}$\uparrow$} & \textbf{Table\textsuperscript{TEDS-S}$\uparrow$} & \textbf{Reading Order\textsuperscript{Edit}$\downarrow$} \\
        \midrule

        \multirow{7}{*}{\textbf{General VLMs}}
        & InternVL3.5-241B~\cite{wang2025internvl35} & 241B & 83.76 & 0.130 & 89.95 & 74.35 & 79.78 & 0.215 \\
        & Kimi K2.5~\cite{kimiteam2026kimik25visualagentic} & 1T & 84.53 & 0.107 & 83.50 & 80.76 & 84.00 & 0.211 \\
        & GPT-5.2~\cite{gpt5_2} & - & 86.59 & 0.114 & 88.21 & 82.95 & 87.93 & 0.193 \\
        & Qwen3-VL-235B~\cite{yang2025qwen3} & 235B & 89.78 & 0.063 & 92.55 & 83.07 & 86.75 & 0.166 \\
        & Gemini 3 Flash~\cite{gemini30} & - & 92.62 & 0.066 & 95.16 & 89.29 & 93.51 & 0.172 \\
        & Gemini 3 Pro~\cite{gemini30} & - & 92.91 & 0.064 & 95.99 & 89.15 & 92.96 & 0.165 \\
        & Ovis2.6-30B-A3B~\cite{lu2024ovisstructuralembeddingalignment,lu2025ovis25technicalreport} & 30B & 93.70 & \cellcolor{cyan!15}\underline{0.035} & 95.17 & 89.44 & 92.40 & 0.135 \\
        \midrule

        \multirow{21}{*}{\textbf{Specialized VLMs}}
        & POINTS-Reader~\cite{liu2025points} & 3B & 83.37 & 0.096 & 85.72 & 73.98 & 77.40 & 0.198 \\
        & Nanonets-OCR-s~\cite{Nanonets-OCR-S} & 3B & 83.61 & 0.108 & 81.46 & 80.18 & 84.51 & 0.213 \\
        & Mistral OCR~\cite{mistral} & - & 85.66 & 0.097 & 89.91 & 76.78 & 80.93 & 0.171 \\
        & olmOCR~\cite{poznanski2025olmocr} & 7B & 85.74 & 0.139 & 88.10 & 83.00 & 87.17 & 0.216 \\
        & Dolphin-1.5~\cite{feng2025dolphin} & 0.3B & 86.52 & 0.094 & 87.49 & 81.43 & 84.82 & 0.167 \\
        & MonkeyOCR-pro-3B~\cite{li2025monkeyocr} & 3B & 88.57 & 0.074 & 88.74 & 84.35 & 88.62 & 0.189 \\
        & OCRVerse~\cite{zhong2026ocrverseholisticocrendtoend} & 4B & 88.60 & 0.063 & 89.61 & 82.44 & 86.27 & 0.163 \\
        & Dolphin-v2~\cite{feng2025dolphin} & 3B & 89.50 & 0.069 & 91.01 & 84.40 & 87.44 & 0.150 \\
        & HunyuanOCR~\cite{hunyuanvisionteam2025hunyuanocrtechnicalreport} & 1B & 89.95 & 0.088 & 87.68 & 91.01 & 93.23 & 0.171 \\
        & DeepSeek-OCR 2~\cite{wei2025deepseek} & 3B & 90.25 & 0.050 & 91.84 & 83.89 & 87.75 & 0.144 \\
        & OpenDoc-0.1B~\cite{du2025unirec01bunifiedtextformula} & 0.1B & 90.67 & 0.049 & 93.02 & 83.88 & 87.45 & 0.140 \\
        & dots.ocr~\cite{dotsocr} & 3B & 90.77 & 0.048 & 89.95 & 87.18 & 90.58 & 0.138 \\
        & MinerU-2.5~\cite{niu2025mineru2} & 1.2B & 93.04 & 0.045 & 95.77 & 87.88 & 91.47 & 0.130 \\
        & FireRed-OCR~\cite{fireredocr} & 2B & 93.26 & 0.037 & 95.44 & 88.04 & 91.06 & 0.131 \\
        & Logics-Parsing-v2~\cite{LogicsParsing} & 4B & 93.33 & 0.041 & 95.65 & 88.42 & 91.98 & 0.137 \\
        & Youtu-Parsing~\cite{YoutuParsing} & 2.5B & 93.74 & 0.044 & 93.63 & 92.02 & 95.00 & \cellcolor{red!15}\textbf{0.116} \\
        & Qianfan-OCR~\cite{QianfanOCR} & 4B & 93.90 & 0.040 & 95.08 & 90.53 & 93.31 & 0.130 \\
        & PaddleOCR-VL~\cite{cui2025paddleocrvl} & 0.9B & 94.18 & 0.040 & 95.91 & 90.65 & 93.74 & 0.135 \\
        & PaddleOCR-VL-1.5~\cite{cui2026paddleocrvl15multitask09bvlm} & 0.9B & 94.93 & 0.038 & 96.89 & 91.67 & 94.37 & 0.130 \\
        & GLM-OCR~\cite{duan2026glmocrtechnicalreport} & 0.9B & 95.22 & 0.044 & 97.18 & 92.83 & 95.39 & 0.133 \\
        & MinerU2.5-Pro~\cite{MinerU25Pro} & 1.2B & \cellcolor{cyan!15}\underline{95.75} & 0.036 & \cellcolor{cyan!15}\underline{97.45} & \cellcolor{cyan!15}\underline{93.42} & \cellcolor{cyan!15}\underline{95.92} & \cellcolor{cyan!15}\underline{0.120} \\
        & PaddleOCR-VL-1.6 & 0.9B & \cellcolor{red!15}\textbf{96.33} & \cellcolor{red!15}\textbf{0.033} & \cellcolor{red!15}\textbf{97.49} & \cellcolor{red!15}\textbf{94.76} & \cellcolor{red!15}\textbf{97.11} & 0.127 \\
        \bottomrule
    \end{tabular}%
    }
    \caption{Comprehensive evaluation on OmniDocBench v1.6. Performance metrics are cited from the official leaderboard~\cite{omni1.6}. PaddleOCR-VL-1.6 achieves the best overall performance among all evaluated models.}
    \label{tab:omni16_full_performance}
\end{table}

\paragraph{Real5-OmniDocBench}

Real5-OmniDocBench~\cite{zhou2026real5omnidocbenchfullscalephysicalreconstruction} is a recently proposed benchmark designed to evaluate document parsing models under real-world conditions. Built upon OmniDocBench v1.5, it covers five representative scenarios: scanning, warping, screen photography, illumination variation, and skew. Except for the scanning subset, all images are manually captured using handheld mobile devices, closely simulating practical document acquisition settings. Each subset maintains a one-to-one correspondence with the original OmniDocBench samples and follows the same ground-truth annotations and evaluation protocols. With its physically acquired and scenario-diverse data, Real5-OmniDocBench provides a rigorous testbed for assessing the robustness of document parsing models in practical applications.

As illustrated in Table~\ref{tab:real5-omnidocbench-performance}, PaddleOCR-VL-1.6 achieves the best overall performance on Real5-OmniDocBench, setting a new state-of-the-art result with an overall score of 93.19\%. Compared with its predecessor PaddleOCR-VL-1.5, it improves the overall score by 1.14 points, from 92.05\% to 93.19\%. Despite its compact 0.9B parameter scale, PaddleOCR-VL-1.6 outperforms substantially larger general-purpose VLMs, including Qwen3-VL-235B and Gemini-3 Pro, highlighting its strong parameter efficiency for document-centric tasks. 

\begin{table}[H]
    \centering
    \resizebox{\textwidth}{!}{%
    \renewcommand{\arraystretch}{1.2}
    \begin{tabular}{l|ll|c|c c c c c}
        \toprule
        \textbf{Model Type} & \textbf{Methods} & \textbf{Parameters} & \textbf{Overall$\uparrow$} & \textbf{Scanning$\uparrow$} & \textbf{Warping$\uparrow$} & \textbf{Screen Photography$\uparrow$} & \textbf{Illumination$\uparrow$} & \textbf{Skew$\uparrow$}\\
        \midrule
        \multirow{2}{*}{\textbf{Pipeline Tools}} 
         & Maker-1.8.2~\cite{vik2024marker} & - & 60.10 & 70.27 & 58.98 & 63.65 & 66.31 & 41.27 \\
         & PP-StructureV3~\cite{cui2025paddleocr} & - & 64.45 & 84.68 & 59.34 & 66.89 & 73.38 & 37.98 \\
         \midrule
         \multirow{5}{*}{\textbf{General VLMs}} 
         & GPT-5.2~\cite{gpt5_2} & - & 78.66 & 84.43 & 76.26 & 76.75 & 80.88 & 75.00 \\
         & Qwen2.5-VL-72B~\cite{bai2025qwen2} & 72B & 86.92 & 86.19 & 87.77 & 86.48 & 87.25 & 86.90 \\
         & Gemini-2.5 Pro~\cite{gemini25} & - & 88.21 & 89.25 & 87.63 & 87.11 & 87.97 & 89.07\\
         & Qwen3-VL-235B-A22B-Instruct~\cite{yang2025qwen3} & 235B & 88.904 & 89.43 & 89.99 & 89.27 & 89.27 & 86.56 \\
         & Gemini-3 Pro~\cite{gemini30} & - & 89.24 & 89.47 & 88.90 & 88.86 & 89.53 & 89.45 \\
         \midrule
         \multirow{16}{*}{\textbf{Specialized VLMs}}
         & Dolphin-1.5~\cite{feng2025dolphin} & 0.3B & 61.48 & 83.39 & 50.50 & 69.76 & 75.61 & 28.16\\
         & Dolphin~\cite{feng2025dolphin} & 0.3B & 61.78 & 72.16 & 60.35 & 64.29 & 67.29 & 44.83\\
         & Deepseek-OCR 2~\cite{wei2025deepseek} & 3B & 73.01 & 89.59 & 66.53 & 71.65 & 76.02 & 61.28\\
         & Deepseek-OCR~\cite{wei2025deepseek} & 3B & 73.99 & 86.17 & 67.20 & 75.31 & 78.10 & 63.01\\
         & MinerU2-VLM~\cite{MinerU2} & 0.9B & 76.95 & 83.60 & 73.73 & 78.77 & 80.51 & 68.16\\
         & MonkeyOCR-pro-1.2B~\cite{li2025monkeyocr} & 1.9B & 77.15 & 84.64 & 76.59 & 80.24 & 82.11 & 62.18\\
         & MonkeyOCR-3B~\cite{li2025monkeyocr} & 3.7B & 78.29 & 84.65 & 77.27 & 80.71 & 83.16 & 65.67\\
         & MonkeyOCR-pro-3B~\cite{li2025monkeyocr} & 3.7B & 79.49 & 86.94 & 78.90 & 82.44 & 84.71 & 64.47\\
         & Nanonets-OCR-s~\cite{Nanonets-OCR-S} & 3B & 84.19 & 85.52 & 83.56 & 84.86 & 85.01 & 81.98\\
         & PaddleOCR-VL~\cite{cui2025paddleocrvl} & 0.9B & 85.54 & 92.11 & 85.97 & 82.54 & 89.61 & 77.47\\
         & MinerU2.5~\cite{niu2025mineru2} & 1.2B & 85.61 & 90.06 & 83.76 & 89.41 & 89.57 & 75.24\\
         & dots.ocr~\cite{dotsocr} & 3B & 86.38 & 86.87 & 86.01 & 87.18 & 87.57 & 84.27\\
         & MinerU2.5-pro~\cite{MinerU25Pro} & 1.2B & 88.96 & 92.11 & 88.72 & 91.29 & 91.42 & 81.26\\
         & GLM-OCR~\cite{duan2026glmocrtechnicalreport} & - & 90.32 & 92.67 & 90.68 & 91.75 & 91.12 & 85.39\\
         & \textbf{PaddleOCR-VL-1.5}~\cite{cui2026paddleocrvl15multitask09bvlm} & 0.9B & \cellcolor{cyan!15}\underline{92.05} & \cellcolor{cyan!15}\underline{93.43} & \cellcolor{cyan!15}\underline{91.25} & \cellcolor{cyan!15}\underline{91.76} & \cellcolor{cyan!15}\underline{92.16} & \cellcolor{cyan!15}\underline{91.66}\\
         & \textbf{PaddleOCR-VL-1.6} & 0.9B & \cellcolor{red!15}\textbf{93.19} & \cellcolor{red!15}\textbf{94.74} & \cellcolor{red!15}\textbf{92.48} & \cellcolor{red!15}\textbf{92.78} & \cellcolor{red!15}\textbf{93.28} & \cellcolor{red!15}\textbf{92.66}\\
        \bottomrule
    \end{tabular}%
    }
    \caption{Comprehensive evaluation of document parsing on Real5-OmniDocBench.}
    \label{tab:real5-omnidocbench-performance}
\end{table}

\subsection{Core Sub-Capabilities}

This section presents a detailed evaluation of PaddleOCR-VL-1.6 across multiple core sub-capabilities, covering hard table recognition, chart parsing, text spotting, and seal recognition.

\subsubsection{Hard Table Recognition}

\paragraph{In-house-Table.}
Our in-house evaluation set contains 1,258 challenging table samples with comprehensive annotations and fine-grained type labels. It covers 20 table categories, including Chinese, English, and mixed Chinese-English tables, as well as tables with full, partial, or no borders. The set further includes diverse table formats and scenarios, such as formula tables, dense tables, book and manual tables, lists, academic papers, merged-cell tables, low-quality scans, watermarked tables, registration forms, statistical forms, research and financial reports, image-based tables, invoices, and handwritten tables.

Table~\ref{tab:hard_table_recognition} compares different methods on the In-house-Table benchmark. PaddleOCR-VL-1.6 achieves the highest scores on both Overall TEDS (91.71) and Structural TEDS (94.67), demonstrating its effectiveness and reliability in challenging table recognition scenarios.

\begin{table}[H]
    \centering
    \scriptsize
    \setlength{\tabcolsep}{4pt}
    \renewcommand{\arraystretch}{0.85}
    \begin{tabular}{l|c c}
        \toprule
        \textbf{Methods} & \textbf{Overall TEDS$\uparrow$} & \textbf{Structural TEDS$\uparrow$} \\
        \midrule
        MonkeyOCR~\cite{li2025monkeyocr} & 73.96 & 78.24 \\
        Qwen2.5-VL-3B~\cite{bai2025qwen2} & 73.98 & 77.65 \\
        dots.ocr~\cite{dotsocr} & 75.47 & 79.14 \\
        Qwen2.5-VL-7B~\cite{bai2025qwen2} & 75.49 & 79.26 \\
        OCRFlux-3B~\cite{OCRFlux} & 77.41 & 80.71 \\
        Qwen2.5-VL-72B~\cite{bai2025qwen2} & 77.62 & 83.61 \\
        Nanonets-OCR-s~\cite{Nanonets-OCR-S} & 78.24 & 81.90 \\
        MinerU2-VLM~\cite{MinerU2} & 82.86 & 87.30 \\
        MinerU2.5~\cite{niu2025mineru2} & 84.69 & 89.55 \\
        TRivia-3B~\cite{TRivia} & 86.12 & 91.16 \\
        GLM-OCR~\cite{duan2026glmocrtechnicalreport} & 86.21 & 90.76 \\
        PaddleOCR-VL~\cite{cui2025paddleocrvl} & 86.99 & 90.66 \\
        PaddleOCR-VL-1.5\cite{cui2026paddleocrvl15multitask09bvlm} & 87.14 & 90.61 \\
        MinerU2.5-Pro\cite{MinerU25Pro} & \cellcolor{cyan!15}\underline{89.77} & \cellcolor{cyan!15}\underline{93.78} \\
        \textbf{PaddleOCR-VL-1.6} & \cellcolor{red!15}\textbf{91.71} & \cellcolor{red!15}\textbf{94.67} \\
        \bottomrule
    \end{tabular}%
    \caption{Evaluation results on the hard table recognition benchmark.}
    \label{tab:hard_table_recognition}
\end{table}

\subsubsection{Chart Parsing}

\paragraph{In-house-Chart.}
Our in-house chart recognition evaluation set contains 1,801 samples, all of which have undergone rigorous manual review to ensure annotation correctness. The set covers 11 chart categories, including bar-line hybrid, pie, 100\% stacked bar, area, bar, bubble, histogram, line, scatterplot, stacked area, and stacked bar charts. It includes 851 English samples and 950 Chinese samples. Before evaluation, both predicted and ground-truth data tables are normalized into a unified Markdown format to reduce expression ambiguity.

As shown in Table~\ref{tab:chart_parsing}, PaddleOCR-VL-1.6 achieves the strongest chart parsing performance on the In-house-Chart benchmark, with RMS-F1\cite{liu2023deplotoneshotvisuallanguage} scores of 91.74 overall, 90.11 on English charts, and 93.37 on Chinese charts. It outperforms its predecessors, PaddleOCR-VL and PP-StructureV3, highlighting its strong ability to recover structured data from complex charts.

\begin{table}[H]
    \centering
    \scriptsize
    \setlength{\tabcolsep}{5pt}
    \renewcommand{\arraystretch}{0.9}
    \begin{tabular}{l|c c c}
        \toprule
        \multirow{2}{*}{\textbf{Models}} & \multicolumn{3}{c}{\textbf{RMS-F1$\uparrow$}} \\
        \cmidrule(lr){2-4}
        & \textbf{Overall} & \textbf{EN} & \textbf{ZH} \\
        \midrule
        TinyChart~\cite{zhang2024tinychart} & 21.59 & 47.26 & 8.76 \\
        GOT~\cite{GOT20} & 31.60 & 11.00 & 41.90 \\
        OneChart~\cite{OneChart} & 37.16 & 13.84 & 48.82 \\
        qwenVL-2.5-72B~\cite{Qwen25VL} & 73.00 & 69.72 & 74.64 \\
        HunyuanOCR~\cite{hunyuanvisionteam2025hunyuanocrtechnicalreport} & 75.13 & 65.54 & 79.92 \\
        PaddleOCR-VL-1.5~\cite{cui2026paddleocrvl15multitask09bvlm} & 80.37 & 76.15 & 84.58 \\
        PP-StructureV3~\cite{cui2025paddleocr} & 80.60 & 79.63 & 81.09 \\
        PaddleOCR-VL~\cite{cui2025paddleocrvl} & 84.40 & 82.22 & 85.49 \\
        Gemini 3 Flash~\cite{gemini30} & \cellcolor{cyan!15}\underline{89.45} & \cellcolor{cyan!15}\underline{88.23} & \cellcolor{cyan!15}\underline{90.66} \\
        \textbf{PaddleOCR-VL-1.6} & \cellcolor{red!15}\textbf{91.74} & \cellcolor{red!15}\textbf{90.11} & \cellcolor{red!15}\textbf{93.37} \\
        \bottomrule
    \end{tabular}
    \caption{Comparison of chart parsing performance on the in-house chart benchmark.}
    \label{tab:chart_parsing}
\end{table}

\subsubsection{Text Spotting}

\paragraph{In-house-Text-Spotting.}
The in-house text spotting benchmark evaluates end-to-end OCR capability, covering both text detection and recognition. It spans 9 representative dimensions, including common scenes, Japanese text, degraded or low-quality images, Chinese and English handwriting, table-structured content, ancient documents, and Traditional Chinese. These categories are designed to reflect diverse document scenarios and practical deployment challenges, ranging from regular printed text to layout-sensitive, low-quality, handwritten, and historically styled materials.

As summarized in Table~\ref{tab:spotting}, PaddleOCR-VL-1.6 achieves the highest spotting accuracy across all 9 evaluated dimensions, consistently outperforming strong baselines. These results demonstrate its robust generalization across diverse visual conditions, text styles, and document layouts, indicating that the model remains reliable in both standard OCR scenarios and challenging real-world settings that require precise localization and faithful transcription.

\begin{table}[htbp]
\centering
\resizebox{\textwidth}{!}{
\begin{tabular}{l|c|cccccccccc}
\toprule
\textbf{Dataset} 
& \textbf{Overall}
& \textbf{Ancient} 
& \textbf{Blur} 
& \textbf{Common} 
& \textbf{\makecell{Handwrite\\\_ch}} 
& \textbf{\makecell{Handwrite\\\_en}} 
& \textbf{\makecell{Printing\\\_ch}} 
& \textbf{\makecell{Printing\\\_en}} 
& \textbf{Table} 
& \textbf{Japanese} \\ 

\midrule

HunyuanOCR~\cite{hunyuanvisionteam2025hunyuanocrtechnicalreport}
& 62.90
& 61.64 & 63.92 & 52.22 & 79.84 & 76.65 & 62.13 & 59.56 & 44.19  & 65.93 \\

Rex-Omni~\cite{jiang2025detect} 
& 66.82
& 42.51 & 69.36 & 61.12 & 81.47 & 78.12 & 69.61 & 60.88 & 71.85  & 66.42\\

\textbf{PaddleOCR-VL-1.5}~\cite{cui2026paddleocrvl15multitask09bvlm}
& 86.21 
& 85.23 & 84.22 & 77.13 & \cellcolor{red!15}\textbf{89.52} & 91.63 & 86.69 & \cellcolor{red!15}\textbf{86.89} & 89.93 & 84.61 \\ 

\textbf{PaddleOCR-VL-1.6} 
& \cellcolor{red!15}\textbf{87.47}
& \cellcolor{red!15}\textbf{85.98} 
& \cellcolor{red!15}\textbf{90.59} 
& \cellcolor{red!15}\textbf{77.28} 
& 85.90
& \cellcolor{red!15}\textbf{92.60} 
& \cellcolor{red!15}\textbf{91.26} 
& 86.51
& \cellcolor{red!15}\textbf{92.32} 
& \cellcolor{red!15}\textbf{84.76} \\

\bottomrule
\end{tabular}
}
\caption{Comparison of text spotting performance on the in-house benchmark. Overall denotes the average accuracy across all 9 evaluation dimensions.}
\label{tab:spotting}
\end{table}

\subsubsection{Seal Recognition}
\label{Seal Recognition}

\paragraph{In-house-Seal.}
The in-house seal recognition benchmark is designed to evaluate model performance on specialized seal text recognition. It contains 300 high-quality images covering diverse seal shapes, including circular, oval, and rectangular seals, as well as challenging real-world conditions such as overlapping text, low-contrast impressions, and distorted backgrounds. Normalized Edit Distance (NED) is used as the primary metric to measure character-level recognition accuracy.

As illustrated in Table~\ref{table:seal_results}, PaddleOCR-VL-1.6 demonstrates a clear advantage in seal recognition. Despite its compact 0.9B parameter scale, it achieves an NED of 0.119, substantially outperforming the 235B-parameter Qwen3-VL with an NED of 0.382, as well as its predecessor. These results highlight the model's effectiveness in handling specialized document elements.

\begin{table}[ht]
\centering
 \fontsize{8}{8}\selectfont
\renewcommand{\arraystretch}{1.2}

\begin{tabular}{l|c|c}
\toprule
\textbf{Model} & \textbf{Parameters} & \textbf{NED ($\downarrow$)} \\ 
\midrule
Qwen2.5-VL-72B~\cite{bai2025qwen2} & 72B & 0.396 \\
Qwen3-VL-235B-A22B-Instruct~\cite{yang2025qwen3} & 235B & 0.382 \\ 
PaddleOCR-VL-1.5~\cite{cui2026paddleocrvl15multitask09bvlm} & 0.9B & 0.138 \\
\textbf{PaddleOCR-VL-1.6} & 0.9B & \cellcolor{red!15}\textbf{0.119} \\ 
\bottomrule
\end{tabular}
\caption{Comparison of seal recognition performance on in-house-seal benchmark.}
\label{table:seal_results}
\end{table}

\subsection{Ablation Study}

We conduct an ablation study on OmniDocBench v1.6 to analyze the contribution of each post-training stage in PaddleOCR-VL-1.6. Starting from the PaddleOCR-VL-1.5 checkpoint, we progressively apply continued pre-training (CPT), supervised fine-tuning (SFT), and reinforcement learning (RL). This evaluation traces how the model evolves across representative parsing metrics, including the overall score, text edit distance, formula CDM, Table-TEDS, and Table-TEDS-S. 

\begin{table}[H]
    \centering
    \scriptsize
    \setlength{\tabcolsep}{5pt}
    \renewcommand{\arraystretch}{0.9}
    \begin{tabular}{l|c c c c c}
        \toprule
        \textbf{Stage} & \textbf{Overall$\uparrow$} & \textbf{Text$^{\mathrm{Edit}}\downarrow$} & \textbf{Formula$^{\mathrm{CDM}}\uparrow$} & \textbf{Table$^{\mathrm{TEDS}}\uparrow$} & \textbf{Table$^{\mathrm{TEDS\text{-}S}}\uparrow$} \\
        \midrule
        PaddleOCR-VL-1.5~\cite{cui2026paddleocrvl15multitask09bvlm} & 94.93 & 0.038 & 96.89 & 91.67 & 94.37 \\
        + CPT & 95.62 & 0.035 & 97.32 & 93.03 & 95.82 \\
        + SFT & 96.25 & 0.034 & 97.37 & 94.74 & 97.09 \\
        + RL & \textbf{96.33} & \textbf{0.033} & \textbf{97.49} & \textbf{94.76} & \textbf{97.11} \\
        \bottomrule
    \end{tabular}
    \caption{Ablation study of the progressive post-training stages on OmniDocBench v1.6.}
    \label{tab:ablation_study}
\end{table}

Table~\ref{tab:ablation_study} reports the contribution of each progressive post-training stage on OmniDocBench v1.6. Starting from PaddleOCR-VL-1.5, the full recipe improves the overall score from 94.93\% to 96.33\%, while consistently improving text recognition, formula recognition, and table recognition metrics. The largest gains come from the CPT and SFT stages. CPT raises the overall score by 0.69 points and substantially improves Table-TEDS from 91.67\% to 93.03\%, showing that broad distributional expansion and corrected supervision from the data engine provide a strong foundation for further optimization. SFT brings another 0.63 points overall improvement and further increases Table-TEDS to 94.74\% and Table-TEDS-S to 97.19\%, indicating that high-quality hard samples are particularly effective for refining hard samples.

The RL stage brings a smaller but still positive gain, further improving the overall score from 96.25\% to 96.33\% and increasing the Formula-CDM score from 97.37\% to 97.49\%. This relatively modest improvement is expected, as the model has already reached a strong performance regime after CPT and SFT on OmniDocBench v1.6, leaving less headroom for additional optimization. Nevertheless, RL further refines the final model through reward-guided training, contributing to the best overall performance. These results suggest that, for document parsing, the major performance gains come from high-quality data construction and staged supervised adaptation, while RL serves as a final refinement step for pushing an already strong model closer to its performance ceiling.

\section{Conclusion}

This work presents PaddleOCR-VL-1.6, an enhanced compact document parsing model that builds upon PaddleOCR-VL-1.5 while preserving its efficient 0.9B architecture. Instead of relying on indiscriminate model scaling, PaddleOCR-VL-1.6 improves performance through an under-optimized-region-driven data engine and a progressive post-training pipeline covering CPT, SFT, and RL. The resulting model achieves state-of-the-art performance on OmniDocBench v1.6 and demonstrates strong robustness on Real5-OmniDocBench, while also delivering consistent gains across key sub-capabilities such as hard table recognition, chart parsing, text spotting, and seal recognition. These results show that targeted data optimization and staged post-training can effectively unlock the remaining potential of compact document VLMs. By providing accurate and robust document understanding across diverse real-world scenarios, PaddleOCR-VL-1.6 offers a high-quality parsing foundation for downstream RAG systems, large language model applications, and practical document intelligence workflows.

\bibliography{main}

\setcounter{figure}{0}
\makeatletter 
\renewcommand{\thefigure}{A\@arabic\c@figure}
\makeatother

\setcounter{table}{0}
\makeatletter 
\renewcommand{\thetable}{A\@arabic\c@table}
\makeatother

\end{document}